\documentclass[runningheads]{llncs}
\pdfoutput=1
\usepackage{graphicx}
\usepackage{amsmath}
\usepackage{amssymb}
\usepackage{url} 
\def\calt{{\mathcal{T}}}

\def\naf{{ not \:}}
\def\hp{{\textnormal -}}

\def\kbbio{{\tt KB\_Bio\_101}}
\def\fdnc{\ensuremath{\mathbb{FDNC}}}
\def\datalogplusminus{\ensuremath{\mathit{Datalog}^\pm}}
\def\aspfs{\ensuremath{\mathit{ASP}^{fs}}}

\allowdisplaybreaks

\titlerunning{Query Answering in Object Oriented Knowledge Bases in Logic Programming\hspace*{-1ex}} 
\authorrunning{V.K.~Chaudhri \emph{et al}\/.}
\title{Query Answering in Object Oriented Knowledge Bases in Logic Programming: 
Description and Challenge for ASP 
} 
\author{
Vinay K. Chaudhri\inst{1}
\and Stijn Heymans\inst{1}
\and Michael Wessel\inst{1} 
\and Tran Cao Son\inst{2}
}
\institute{
Artificial Intelligence Center, SRI International, Menlo Park, CA  94025, USA
\and 
Computer Science Department, New Mexico State University, NM 88003, USA 
}

\begin{document}
\maketitle

\begin{abstract}
Research on developing efficient and scalable ASP solvers can
substantially benefit by the availability of data sets to experiment
with.  \kbbio\ contains knowledge from a biology textbook, has been
developed as part of Project Halo, and has recently become available
for research use.  \kbbio\ is one of the largest KBs available in ASP
and the reasoning with it is undecidable in general.  We give a
description of this KB and ASP programs for a suite of queries that
have been of practical interest. We explain why these queries pose
significant practical challenges for the current ASP solvers.
\end{abstract}

\section{Introduction}

The \kbbio\ represents knowledge from a textbook used for advanced
high school and introductory college biology courses \cite{Reece+11}.
The KB was developed by SRI as part of their work for Project
Halo\footnote{\url{http://www.projecthalo.com/}} and contains a
concept taxonomy for the whole textbook and detailed rules for 20
chapters of the textbook.  SRI has tested the educational usefulness
of this KB in the context of an intelligent textbook called {\em
  Inquire}\footnote{\url{http://www.aaaivideos.org/2012/inquire_intelligent_textbook/}}.

The {\tt KB\_Bio\_101} was originally developed using a knowledge
representation and reasoning system called Knowledge Machine (KM)
\cite{ClarkP06}.  To express {\tt KB\_Bio\_101} in answer set
programming (ASP) required us to define a conceptual modeling layer
called {\em Object Oriented Knowledge Base} or OOKB
\cite{ChaudhriHMS12a}.  The goal of this paper is not to introduce
OOKB as a more complete specification and analysis of formal
properties of OOKBs are available elsewhere~\cite{ChaudhriHMS12a}.
OOKB is of more general interest as it supports conceptual modeling
primitives that are commonly found in description logic (DL) family of
languages such as a facility to define classes and organize them into
a hierarchy, define partitions, ability to define relations (also
known as slots) and organize them into a relation hierarchy, support
for domain, range and qualified number constraints, support for
defining sufficient conditions of a class, and support for descriptive
rules.  The features in OOKB also overlap with the features of logic
programming (LP) languages such as \fdnc\ \cite{eiter2010},
\datalogplusminus\ \cite{cali2009}, and \aspfs\ \cite{AlvianoFL10} in
its support for function symbols. It differs from these previous LP
languages as well as from the DL systems in that the functions can be
used to specify graph-structured objects which cannot be done in these
other languages.  The reasoning with OOKBs has been proven to be
undecidable \cite{ChaudhriHMS12a}.  

The approach taken in this paper fosters work on multi-paradigm problem solving
in the following ways.  First, it aims to give a declarative formalization of
reasoning tasks that were originally implemented in KM which is a very 
different paradigm for reasoning as compared to ASP.  Second, the conceptual
modeling primitives considered here directly overlap with many description logics,
thus, providing another example of integration between ASP with DLs.

The primary objective of this paper is to introduce\kbbio as a
valuable and data set and four queries of practical interest on this
KB.  These queries have been found extremely useful in the context of
{\em Inquire}. This dataset presents an excellent opportunity for
further development of ASP solvers for the following reasons.
\begin{list}{$\bullet$}{\itemsep=0pt \topsep=0pt \parsep=0pt \leftmargin=10pt} 
\item
Recent developments in ASP suggest that it could potentially
provide an ideal tool for large scale KBs. Yet, most of the KBs
described in the literature are fairly small. \kbbio\ 
provides a real-world ASP program that fits this bill.

\item We note that \kbbio\ contains rules with function symbols for
  which the grounding is infinite. A simple example
  is a KB consisting of a single class {\tt person}, and a single
  relation {\tt has-parent}, and a statement of the form ``for each
  {\tt person} there exists an instance of the {\tt has-parent}
  relation between this person with another individual who is also a
  {\tt person}''. The skolemized versions of these statements require
  function symbols.  An obvious first challenge that must be addressed
  is to develop suitable grounding techniques.

\item Even though rules in \kbbio\ follow a small number of axiom
  templates, the size of this KB indicates that this could be a
  non-trivial task for ASP solvers.

\item The \kbbio\ cannot be expressed in commonly available decidable
  DLs because it contains graph structured descriptions. Efficient
  reasoning with graph structures is an area of active recent research
  \cite{MagkaMH12,MotikGHS09}, and since there exists an export of
  \kbbio\ for DL systems also \cite{ORE}, it provides an ideal usecase
  to explore the relative effectiveness of DL reasoners vs ASP solvers
  on a common problem.

\item The reasoning tasks of computing differences between two
  concepts and finding relationships between two individuals are
  computationally intensive tasks. The implementations of these tasks
  in {\em Inquire} rely on graph algorithms and trade completeness for
  efficiency. These tasks will present a tough challenges to ASP
  solvers.

\item Last but not the least, we believe that the KB could entice the
  development and/or experimentation with new solvers for extended
  classes of logic programs (e.g., language with existential
  quantifiers or function symbols).

\end{list}

In addition to the challenges listed above, it will be possible to
define multiple new challenges of increasing difficulty that can be
used to motivate further research and development of ASP solvers.
 
\section{Background: Logic Programming and OOKB} 

\subsection{Logic Programming} 

A logic program $\Pi$ is a set of rules of the form %
\begin{eqnarray}
\label{lprule1}
c  \leftarrow a_1,\ldots,a_m,\naf a_{m+1},\ldots,\naf a_n 
\end{eqnarray}
where $0 {\le} m {\le} n$, each $a_i$  
is a literal of a first order language 
and $\naf a_j$, $m {< } j {\le} n$, is called 
a negation as failure literal (or
naf-literal). $c$ can be a literal or omitted. 
A rule (program) is non-ground if it contains some variable; 
otherwise, it is a ground rule (program). 
When $n = 0$, 
the rule is called a {\em fact}. When $c$ is omitted, the rule is a {\em constraint}. Well-known notions
such as substitution, the Herbrand universe ${\cal U}_\Pi$, and Herbrand base 
${\cal B}_\Pi$ of a program $\Pi$ are defined as usual. 

 \iffalse  

The Herbrand universe of a program $\Pi$
is the set ${\cal U}_\Pi$ of 
terms constructable from constants 
and function symbols in $\Pi$.
The Herbrand base of $\Pi$, ${\cal B}_\Pi$, is the set of 
ground atoms constructable using the predicate symbols in $\Pi$ and the 
terms in ${\cal U}_\Pi$. 
A substitution  $\delta$ is given by a set 
$\{X_1/t_1,\ldots,X_s/t_s\}$ where 
$X_i$'s are distinctive variables and $t_i$'s are terms. 
%$t_i$ is called the value of $X_i$ in $\delta$. 
$\delta$ is a ground substitution if $t_i \in {\cal U}_\Pi$ 
for every $i$.
For a literal $l$, $l[\delta]$ is the literal obtained from $l$ by
simultaneously replacing every occurrence of $X_i$ by $t_i$ 
for every $i$.
 \fi

The semantics of a program is defined over ground programs. 
For a ground rule $r$ of the form (\ref{lprule1}), let 
$pos(r) {=} \{a_1,\ldots,a_m\}$ and 
$neg(r) {=} \{a_{m+1},\ldots, a_n\}$.  
A set of ground literals $X$ is consistent if there exists 
no atom $a$ s.t.  
 $\{a, \neg a\} {\subseteq}  X$. 
 A ground rule $r$ is
{\em satisfied} by $X$ if 
({\em i}) $neg(r) {\cap} X {\ne} \emptyset$; ({\em ii}) 
$pos(r){\setminus} X {\ne} \emptyset$; or ({\em iii})  $c \in X$. 
 
Let $\Pi$ be a ground program.
For a consistent set of ground literals $S$, the {\em reduct} 
of $\Pi$ w.r.t. $S$, denoted by $\Pi^S$, is the program
obtained from the set of all rules of $\Pi$ by deleting
({\bf i}) each rule that has a naf-literal {\em not} $a$ in its body with
$a \in S$, and
({\bf ii}) all naf-literals in the bodies of the remaining rules.
$S$ is an \emph{answer set} of $\Pi$ \cite{GelfondL90} if it
satisfies the following conditions: 
({\bf i})  If $\Pi$ does not contain any naf-literal 
then $S$ is the minimal set of ground literals satisfying 
all rules in $\Pi$; and 
% least fixpoint of the {\em immediate consequence operator}, 
% denoted by $T_\Pi$, that maps sets of ground literals into 
% sets of ground literals;  and
({\bf ii}) If $\Pi$ contains some naf-literal  then $S$ is an answer set of
$\Pi$ if $S$ is the answer set of $\Pi^S$. 

For a non-ground program $\Pi$, a set of 
literals  in ${\cal B}_\Pi$ is an answer set of $\Pi$ if it is an answer set 
of $ground(\Pi)$ that is the set of all possible ground rules obtained from instantiating 
variables with terms in ${\cal U}_\Pi$. $\Pi$ is \emph{consistent} if it has an answer set.  
%$SM(\Pi) \ne \emptyset$ where $SM(\Pi)$ denotes the set of answer sets of $\Pi$. 
$\Pi$ {\em entails} a ground literal 
$a$, $\Pi \models a$, 
if $a$ belongs to every answer set of $\Pi$.
%if $\forall S \in SM(\Pi).[a \in S]$.

For convenience in notation, we will make use of choice atoms 
as defined in \cite{SimonsNS02} that can occur in a rule 
wherever a literal can. 
Answer sets of logic programs can be computed using 
answer set solvers (e.g., {\sc Clasp} \cite{GebserKNS07}, {\bf dlv} \cite{dlv97}).

\iffalse 
A 
choice atom is of the form
$l \: S \: u$ where $S$ is a set of literals and $l \le u$ are
non-negative integers; $l \: S \: u$  is true in a set of literals $X$
if   $l \le | S \cap X| \le u$. When $l=0$ or $u=\infty$, they will be 
omitted.
The set $S$ in a choice atom $l \: S \: u$ can occur in various forms 
(see, e.g.,  \cite{SimonsNS02}); e.g., it can be explicitly listed as 
$\{l_1,\ldots,l_n\}$ where $l_i$'s are literals; or written in the form 
$\{p  : q \}$ where $p$, $q$ are atoms.  
Given a set of ground literal $X$, 
$\{p  : q \} \cap X$ is the set of atoms 
$\{p[\delta] \mid $ there exists a ground instantiation $\delta$ 
such that $q[\delta] \in X\}$. 
Answer sets of logic programs can be computed using 
answer set solvers (e.g., {\sc Clasp} \cite{GebserKNS07}, {\bf dlv} \cite{dlv97}).
\fi
    
\subsection{Object-Oriented Knowledge Bases} 

We will now review the notion of an OOKB \cite{ChaudhriHMS12a}. We note that an OOKB 
could be viewed as a logic program with function symbols and 
the language of OOKBs contains features that cannot be represented in 
previous investigated classes of function symbols such as  
\fdnc\ \cite{eiter2010}, \datalogplusminus\ \cite{cali2009}, or \aspfs\ \cite{AlvianoFL10}. 
In essense, an OOKBs is a logic program consisting of the following components: 

\begin{list}{$\bullet$}{\topsep=0pt \parsep=0pt \itemsep=0pt}  

\item {\em Taxonomic Knowledge:} This group of facts encodes the class hierarchy, 
the relation hierarchy, individual constants and their class membership. It contains  
ASP-atoms of the following form:  \\
{\small
\begin{minipage}[bht]{0.45\linewidth} 
\begin{eqnarray} 
\!\!\!\!&& \!\!\!\!\!\!\!\!\!\!\!\! class(c)    \label{eqclass} \\
\!\!\!\!&& \!\!\!\!\!\!\!\!\!\!\!\! individual(i)    \label{eqconstant} \\
\!\!\!\!&& \!\!\!\!\!\!\!\!\!\!\!\! subclass\_of(c_1, c_2) \quad\:\:\label{eqsubclass}  \\
\!\!\!\!&& \!\!\!\!\!\!\!\!\!\!\!\! disjoint(c_1, c_2)  \label{eqdisjoint}  \\
\!\!\!\!&& \!\!\!\!\!\!\!\!\!\!\!\! instance\_of(i, c)  \label{eqinstance}   
\end{eqnarray}
\end{minipage} 
\begin{minipage}[bht]{0.50\linewidth} 
\begin{eqnarray} 
&&\!\!\!\!\!\!\!\!  relation(r)  \label{sdec0} \\
&&\!\!\!\!\!\!\!\! range(r,c)      \label{sdec1}  \\
&&\!\!\!\!\!\!\!\! domain(r,c)     \label{sdec2} \\
&&\!\!\!\!\!\!\!\! subrelation\_of(r_1, r_2)  \quad \quad \label{sc1}  \\
&&\!\!\!\!\!\!\!\! compose(r_1, r_2, r_3)  \label{sc2} \\
&&\!\!\!\!\!\!\!\! inverse(r_1, r_2)   \label{sc3}  
\end{eqnarray} 
\end{minipage} 
}

\medskip\noindent
The predicate names are self-explanatory. 
  
\item {\em Descriptive statements:}  Relationships between 
individuals are encoded in OOKB by descriptive statements of the form: 
{\small
\begin{eqnarray}
value(r, f(X), g(X))  & \leftarrow & instance\_of(X, c)    \label{eqslot1} \\  
value(r, X, g(X))  & \leftarrow & instance\_of(X, c)    \label{eqslot2}   
\end{eqnarray}  
}

where $f$ and $g$ are unary functions, called {\em Skolem functions},
such that $f \ne g$ and $c$ is a class.  %A statement of the form
Axiom~\ref{eqslot1} (or \ref{eqslot2}) describes a relation value of
individuals belonging to class $c$, encoded by the atom $value(r,
f(X), g(X))$ (or $value(r, X, f(X)$).  It states that for each
individual $X$ in $c$, $f(X)$ (or $X$) is related to $g(X)$ via the
relation $r$. An example use of axiom~\ref{eqslot2} is: Every
Eukaryotic Cell has part a Nucleus, where {\em Eukaryotic Cell} and
{\em Nucleus} are individuals from these two classes, and {\em has
  part} is a relationship between those individuals.  It is required
that if $f$ (or $g$) appears in \eqref{eqslot1} or \eqref{eqslot2},
then the OOKB also contains the following rule

{\small
\begin{eqnarray}
instance\_of(f(X), c_f)   & \leftarrow & instance\_of(X, c) \quad \textnormal{ or }  \\  
instance\_of(g(X), c_g) & \leftarrow & instance\_of(X, c)    
\end{eqnarray}  
}
which specify the class of which $f(X)$ (resp. $g(X)$) 
is a member. For example, if $f(X)$ represents a nucleus individual, then 
$c_f$ will be the class Nucleus.

\item {\em Cardinality constraints on relations:} OOKB allows
  cardinality constraints on relations to be specified by statements
  of the following form: {\small
\begin{eqnarray} 
constraint(t, f(X), r, d, n)    \leftarrow  instance\_of(X, c)    \label{constr} 
\end{eqnarray} 
}
where  $r$ is a relation, $n$ is a non-negative integer, 
$d$ and $c$ are classes, and 
$t$ can either be {\em min}, {\em max}, or {\em exact}. 
This constraint states that 
for each instance $X$ of the class $c$, the set of values 
of relation $r$ restricted on $f(X)$---which  
must occur in a relation value literal $value(r, f(X), g(X))$ 
of $c$---has minimal (resp. maximal, exactly) $n$ values 
belonging to the class $d$. The head of \eqref{constr} is called a constraint literal 
of $c$. 
  
\item {\em Sufficient conditions:} 
A {\em sufficient condition} of a class $c$ defines sufficient
conditions for membership of that class based on the relation values
and constraints applicable to an instance:
{%\small 
\begin{eqnarray} 
instance\_of(X, c) \leftarrow Body(\vec{X}) \label{suffcond} 
\end{eqnarray} 
}
where $Body(\vec{X})$ is a conjunction of relation value literals, 
instance-of literals, constraint-literals  of $c$, 
and $X$ is a variable occurring in the body of the rule.

\item {\em (In)Equality between individual terms:} The rules in this group 
specify in/equality between terms, which are constructable 
from Skolem functions and the variable $X$ ($t_1$ and $t_2$),  
 and have the followimg form: 
\begin{eqnarray} 
eq(t_1, t_2) & \leftarrow &   instance\_of(X, c) \label{unify} \\
neq(t_1, t_2) & \leftarrow &   instance\_of(X, c) \label{unify1}
\end{eqnarray} 
\item {\em Domain-independent axioms:} An OOKB also contains a
set of domain-independent axioms $\Pi_R$ 
for inheritance reasoning, % (rules \eqref{p1}---\eqref{p4}),
reasoning about the relation values of individuals 
 (rules \eqref{p5}---\eqref{p7}),
in/equality between terms
(rules \eqref{r-substitute0}---\eqref{lp-eq9}), and 
enforcing constraints (rules \eqref{constr1}---\eqref{constr6}). 
%This group of rules is omitted for space reason. 
%It can be found in \cite{ChaudhriHMS12a}. 
%It is included in the appendix for reviewing purpose.  
{\small
\begin{eqnarray}  
  subclass\_of(C, B) &\leftarrow &   subclass\_of(C, A),  
   subclass\_of(A, B). \label{p1} \\ %% \nonumber \\
 instance\_of(X, C)  & \leftarrow  & instance\_of(X, D),  
   subclass\_of(D, C).   \label{p2} \\   %% \nonumber \\
   disjoint(C, D)  & \leftarrow &   disjoint(D, C).  \label{p3} \\
    \neg instance\_of(X, C)  & \leftarrow &  instance\_of(X, D),  
disjoint(D, C). \label{p4}   \\
 value(U, X, Z)  & \leftarrow &   compose(S, T, U),     
value(S, X, Y),value(T, Y,Z). \label{p5} \\   
 value(T, X, Y)  & \leftarrow &   subrelation\_of(S, T), value(S, X, Y). \label{p6} \\     
value(T, Y, X)  & \leftarrow &   inverse(S, T), value(S , X, Y).   \label{p7}  \\
 eq(X, Y)  & \leftarrow &  eq(Y, X) \label{r-substitute0} \\
 eq(X, Z)  & \leftarrow &   eq(X, Y), eq(Y, Z),  X \ne Z    \label{r-substitute1}  \\
& \leftarrow & eq(X, Y), neq(X, Y) \label{r-substitute2}   \\
 \{substitute(X, Y)\} & \leftarrow &   eq(X, Y). \label{lp-eq1} \\
& \leftarrow &  eq(X,Y),  \{substitute(X, Z) : eq(X,Z)\} 0, \label{lp-eq2} \\
&& \{substitute(Y, Z) : eq(Y,Z)\} 0. \quad\quad   \nonumber  \\
& \leftarrow &  substitute(X, Y), substitute(X, Z), \label{lp-eq3}  \\
& & X \ne Y, X \ne  Z, Y \ne Z .  \\
& \leftarrow &  substitute(X, Y), X \ne Y, neq(X, Y). \label{lp-eq31}  \\
 substitute(Y, Z)  & \leftarrow &  substitute(X, Z),   X \ne Z,  eq(X, Y). \label{lp-eq4} \\
is\_substituted(X) & \leftarrow &  substitute(X, Y), X\ne Y. \label{lp-eq5}  \\
substitute(X, X) & \leftarrow &    term(X), \naf is\_substituted(X). \label{lp-eq6} \\
term(X)& \leftarrow &  value(S, X, Y). \label{lp-eq8} \\
term(Y)& \leftarrow &  value(S, X, Y). \label{lp-eq9}\\
value_e(S, P, Q) & \leftarrow &  value(S, X, Y), substitute(X, P), substitute(Y, Q). \label{lp-eq7}  \\
  & \leftarrow & value(S, X, Y), domain(S, C),\naf instance\_of(X, C).                \label{constr1} \quad \:\\   
 & \leftarrow  & value(S, X, Y), range(S, C), \naf instance\_of(Y, C).  \quad \label{constr2}   \\ 
 & \leftarrow & constraint(min, Y, S, D, M),  \label{constr3} \\
 &&  \{value_e(S, Y, Z) : instance\_of(Z, D)\} \: M - 1. \nonumber    \\
 & \leftarrow & constraint(max, Y, S, D, M),  \quad\quad  \label{constr4}\\  &&M+1\{value_e(S, Y, Z) : instance\_of(Z, D)\}.   \nonumber \\
 & \leftarrow & constraint(exact, Y, S, D, M), \quad\quad  \label{constr5} \\  && \{value_e(S, Y, Z) : instance\_of(Z, D)\} \: M - 1.  \nonumber \\ 
 & \leftarrow & constraint(exact, Y, S, D, M),  \label{constr6}   \\
 && M+1 \{value_e(S, Y, Z) : instance\_of(Z, D)\}. \quad\quad  \nonumber 
\end{eqnarray}
}
\end{list} 

For a detailed explanation of the above rules, please refer to \cite{ChaudhriHMS12a}.
An {\em OO-domain} is a collection of rules of the form 
\eqref{eqclass}---\eqref{unify1}. 
>From now on, whenever we refer to an OOKB, we mean 
the prorgram $D \cup \Pi_R$, denoted by 
$KB(D)$, where $D$ is the OO-domain of the OOKB\footnote{
In \cite{ChaudhriHMS12a}, general OOKBs, that can contain arbitrary 
logic programming rules, were defined. The discussion
in this paper is applicable to general OOKBs as well. 
}.

\subsection{\kbbio: An OOKB Usage and Some Key Characteristics}

\begin{figure}[!htp]
\centering
 \includegraphics[width=.65\textwidth]{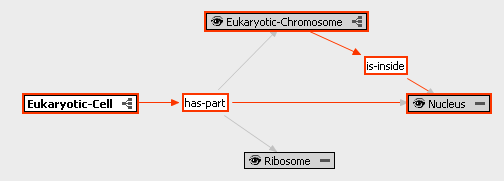}
 \caption{Example graph for ``Eukaryotic-Cell"}\label{fig:cmap}
\end{figure}

The \kbbio\ is an instance of OOKB and is available in ASP format\footnote{See {\tt
    http://www.ai.sri.com/\url{~h}alo/public/exported-kb/biokb.html}}.  The
KB is based on an upper ontology called the Component Library
\cite{Barker}.  The biologists used a knowledge authoring system
called AURA to represent knowledge from a biology textbook.  As an
example, in Figure \ref{fig:cmap}, we show an example AURA
graph. The white node labeled as {\small {\tt Eukaryotic-Cell}} is the
root node and represents the universally quantified variable $X$,
whereas the other nodes shown in gray represent existentials, or the
Skolem functions $f_n(X)$.  The nodes labeled as {\small {\tt
    has\_part}} and {\small {\tt is\_inside}} represent the relation
names.
The authoring process in AURA can be abstractly characterized as
involving three steps: \emph{inherit, specialize and extend.}  For
example, the biologist creates the class {\small\tt  Eukaryotic{\hp}Cell} as a
subclass of {\small\tt  Cell.} While doing so, the system would first
inherit the relation values defined for  {\small\tt  Cell} which in this case
is a  {\small\tt  Chromosome}, and show it in the graphical editor. The
biologist then uses a gesture in the editor to specialize the
inherited  {\small\tt  Chromosome} to a  {\small\tt  Eukaryotic-Chromosome,} and
then introduces a new  {\small\tt   Nucleus} and relates it to the
 {\small\tt  Eukaryotic-Chromosome,} via an  {\small\tt  is{\hp}inside} relationship.
The inherited  {\small\tt  Chromosome} value for the  {\small\tt  has{\hp}part}
relationship, is thus, specialized to  {\small\tt   Eukaryotic{\hp}Chromosome} and
extended by connecting it to the  {\small\tt  Nucleus} by using an
 {\small\tt  is{\hp}inside} relationship.

The statistics about the size of the exported OOKB are summarized in 
Table \ref{table-asp}.  
In total \kbbio\ has more than 300,000 non-ground rules. 
It contains 746 individuals which are members of classes 
which represent constants of measurements, colors, shapes, quantity, 
priority, etc. The KB does not contain individuals 
of biology classes such as {\small \tt cell}, {\small \tt ribosome}, etc.  
For computing properties of an individual or comparing individuals, 
the input needs to introduce the individuals. 
\begin{table}[!htp]
\begin{center} 
\begin{tabular}{|lr||lr|}
\hline
classes & 6430 & $\mathit{domain}$ constraints & 449 \\
individuals & 746 & $\mathit{range}$ constraints & 447 \\
relations& 455 & $\mathit{inverse}$ relation statements & 442 \\
$\mathit{subclass\_of}$ statements & 6993 & $\mathit{compose}$ statements &  431\\
$\mathit{subrelation\_of}$ statements & 297 & qualified number constraints & 936 \\
$\mathit{instance\_of}$ statements & 714 & sufficient conditions & 198 \\
$\mathit{disjoint}$-ness statements & 18616 & descriptive rules & 6430 \\
avg. number of Skolem functions \quad \quad &  24 & equality statements & 108755 \\
in each descriptive rule  & & & \\
\hline
\end{tabular}
\caption{Statistics on \kbbio\label{table-asp}}
\end{center} 
\end{table}
%
\iffalse 
Reasoning with \kbbio\ is generally undecidable as it can contain cyclic representation. Consider that 
we wish to represent the statement: Every {\small\tt biomembrane} has a part
{\small\tt peripheral protein}.  This will be stated as: 
{\small
\begin{equation}  
\begin{array}{rcl} 
instance\_of(f_1(X), peripheral\_protein) &  \leftarrow & instance\_of(X, biomembrane).  \\
value(has\_part, X, f_1(X)) &  \leftarrow & instance\_of(X, biomembrane). 
\end{array} 
\end{equation} 
}
Next, suppose we wish to represent ``Every {\small\tt peripheral protein} is found 
at the outside surface of a {\small\tt biomembrane}''.  This will be stated as:
{\small 
\begin{equation} 
\begin{array}{rcl} 
instance\_of(f_2(X), biomembrane) 
& \leftarrow & instance\_of(X, peripheral\_protein). \\
value(is\_outside, X, f_2(X)) 
& \leftarrow & instance\_of(X, peripheral\_protein). % \nonumber
\end{array} 
\end{equation} 
} 
\noindent
This example leads to a cycle in the representation. The biologists
have been unwilling to give up expressiveness needed to model such an
example.  The current  \kbbio\ has 8604 cycles.

\fi

\iffalse 
\begin{itemize} 
\item a taxonomical knowledge base over a signature with more than 
6000 classes, 450 relations, 800 relation hierarchy statements, 6000 
subclass\_of statements; 
\item about 107000 rules for speficying the relation values of instances of the 6000 classes; and 
\item about 115000 equality specifications; 
\item about ....
\end{itemize}
\fi
 
\section{Queries in OOKBs} 
\label{sec:queries} 
We will now describe the queries given an OOKB, say $KB(D)$. These
queries play a central role in the educational application {\small \tt
  Inquire} \cite{Overholtzer+12} which employs the knowledge encoded
in \kbbio. These queries were developed by extensive analysis of the
questions from an exam, the questions at the back of the book, and the
questions that are educationally useful~\cite{Clark-2003,Chaudhri-13}.

We divide these queries into four groups. The first type of queries
which includes the first two queries asks about facts and
relatiolnships. The second type of queries asks about the taxonomic
information. These first two question types are usually referred to as
the {\em wh-}questions. The third type is about the differences and
similarities between individuals from different classes.  This type of
query has been traditionally studied as an example of analogical
reasoning~\cite{nicholson-2002}.  The fourth type of queries that
includes the last two questions query for relationships between
concepts and are unique to our work.

\begin{list}{$\bullet$}{\itemsep=0pt \parsep=0pt \topsep=0pt}  
\item what is a {\small \tt eukaryotic cell}? 
\item what process provides raw materials for the {\small \tt citric acic cycle} 
during {\em cellular respiration}? 
\item is {\small \tt oocyte} a subclass of a {\small \tt eukaryotic cell}?
\item describe the differences and similarities between 
{\small \tt mitochondrions} and {\small \tt chloroplasts}
\item What is the relationship between a 
{\small \tt mitochondrion} and a {\small \tt chloroplast}
\item in the absence of {\small \tt oxygen}, {\small \tt yeast cells} can obtain {\small \tt energy} by{\em which process}?
\end{list} 

%
%%%  
%To formulate the different types of queries, we need some additional notations.   
%
Let $Z$ be a set of literals of $KB(D)$,
$r$ be a relation, and $i$ be an individual from a class $c$. 
$\calt(i)$ denotes the set of terms constructable 
from Skolem functions and the individual $i$.  
We characterize the set of pairs in the relation $r$ w.r.t. $Z$ 
in $KB(D)$ by the set 
$V(r, i, c, Z) = \{(r,x,y) \mid value(r, x, y) \in Z, x, y \in \calt(i)\}$ 
if  $instance\_of(i, c) \in Z$; otherwise, $V(r, i, c, Z) {=} \emptyset$. 
\begin{definition} 
[Value set of an individual] 
\label{def-sv}
Let $KB(D)$ be an OOKB. For an answer set $M$ of $KB(D)$, 
the {\em value set of an individual $i$ at a class $c$} w.r.t. $M$, $\Sigma(i,c,M)$,  
is defined by $\Sigma(i,c,M) = \bigcup_{relation(r) \in M}  V(i,c,r,M)$.
\end{definition} 
Observe that the rules \eqref{r-substitute1}---\eqref{lp-eq7} 
indicate that $KB(D)$ can have multiple answer sets. Nevertheless, 
the structure of $KB(D)$ allows us to prove the following 
important property of answer sets of $KB(D)$. 
\begin{proposition} \label{prop1} 
Let $KB(D)$ be an OOKB. 
For every two answer sets $M_1$ and $M_2$ of $KB(D)$, 
every literal in $M_1 \setminus M_2$ has one of the 
following forms: ({\em i}) $substitute(x, y)$;
({\em ii}) $is\_substituted(x, y)$; or ({\em iii}) $value_e(r, x, y)$. 
\end{proposition} 
The above proposition indicates that $\Sigma(i,c,M_1) = \Sigma(i,c,M_2)$ 
for arbitrary individual $i$ and class $c$ and 
answer sets $M_1$ and $M_2$ of $KB(D)$. 
The relationship between atoms of the form 
$value(r, x, y)$ and $value_e(r,x,y)$ is as follows. 
\begin{proposition} 
Let $KB(D)$ be an OOKB, $i$ an individual, and $c$ a class. 
For every answer sets $M$ of $KB(D)$, we have that 
 $value_e(r, x, y) \in M$ iff there exists $x', y'$ 
such that ({\em i}) $M$ contains the following atoms 
$eq(x', x)$, $eq(y', y)$, $substitute(x', x)$, and 
$substitute(y', y)$; and ({\em ii}) 
$(r, x', y') \in \Sigma(i,c,M)$.   
\end{proposition} 
The significance of these two propositions is that cautious
reasoning about values of individuals at classes can be accomplished by 
computing {\em one} answer set of $KB(D)$. As we will see, 
the majority of queries is related to this type of reasoning.  

We next describe, for each query $Q$, an input program $I(Q)$ and
a set $R(Q)$ of rules for computing the answer of $Q$. 
Throughout the section, $KB$ denotes an arbitrary but fixed 
OOKB $KB(D)$ and $KB(Q) = KB(D) \cup I(Q) \cup R(Q)$. % for a query $Q$.

\subsection{Subsumption Between Classes ($\mathbf{Q_1}$)} 

Subsumption requires us to compute whether a class $c_1$ 
is subsumed by a class $c_2$, i.e., whether for each answer set $M$ of $KB(Q_1)$, we have for each $instance\_of(x,c_1)\in
M$ also $instance\_of(x,c_2)\in M$. We can answer this question
by introducing  a
fresh constant $i$ in the OOKB and 
set $I(Q_1) = \{instance\_of(i, c_1)\}$.
$R(Q_1)$ consists of a rule:
{ 
\begin{align}
subclass\_of(c_1,c_2) \leftarrow %instance\_of(i, c_1),
                  instance\_of(i,c_2)   \label{q1-rule1}
\end{align} 
}
Indeed, we then have that a class $c_1$ is subsumbed by $c_2$ iff for each
answer set $M$ of $KB(Q_1)$, $subclass\_of(c_1, c_2) \in M$.  
\iffalse 
This conclusion comes from the following observations: 
({\em a}) if $subclass\_of(c_1,c_2)$ can be proved without \eqref{q1-rule1} 
then the subsumption is specified by the class hierarchy of the OO-domain; 
({\em b}) if $subclass\_of(c_1,c_2)$ cannot be proved without \eqref{q1-rule1}, it requires that 
$instance\_of(i,c_2) \in M$, i.e., each instance $i$ 
of $c_1$ is also an instance of $c_2$. 
\fi
Proposition~\ref{prop1} can be extended to $KB(Q_1)$ and thus 
we only need to compute one answer set of $KB(Q_1)$. 
Note that this shows how, as
in description logics, subsumption can be reduced to entailment in the OOKB framework.
We can show that 
\begin{proposition} 
If $KB(Q_1)$ has an answer set $M$ and $subclass\_of(c_1,c_2) \in M$ 
then $c_1$ is subsumed by $c_2$.  
\end{proposition}  
We note that computing answer sets of $KB(Q_1)$ is not a simple 
task (see \cite{ChaudhriHMS12a}). In particular, the problem for 
\kbbio\ is quite challenging due to its size and the potential 
infiniteness of the grounding program of $KB(Q_1)$.   

One can define many more taxonomic queries than what we have
considered here.  Some examples of such queries are as follows. Given
a class $C$, compute all its super classes or subclasses? Given a class,
return only most specific superclass?  Given two classes, return there
nearest common superclass?

Some of the taxonomic queries can require a higher order representation.
For example, given two classes, compute a class description that is
their union or intersection. Such queries are straightforward in a DL
system, and are examples of capabilities that are challenging for the
current ASP systems.
 
\subsection{Description of an Individual ($\mathbf{Q_2}$)}

Queries about the description of an individual ask for a description of 
an individual of a class $c$, represented by a fresh constant  
$i$ in the language of $KB(D)$. This query can be represented by the    
program $I(Q_2) = \{get\_value(i,c). instance\_of(i,c). \}$ 
where $get\_value(i,c)$ encodes the query of ``inquiring about values   
of $i$ at the class $c$.''
We will now discuss the answer to this query.  Intuitively,   
a complete description of $i$ should contain the following information:
\begin{list}{$\bullet$}{\itemsep=0pt \parsep=0pt \topsep=0pt} 
\item $C(c) {=} \{d \mid KB(D) {\models} subclass\_of(c, d)\}$, 
the classes from which $i$ inherits its relation values; and

\item its relation values, i.e., the triples in $\Sigma(i, c, M)$ where
$M$ is a given answer set of $R(Q_2)$.
\end{list}  
Computing a complete description of $i$ could be achieved 
by the following rules:
{\small
\begin{eqnarray} 
out\_member\_of(Y) & \leftarrow & get\_value(I, C),  instance\_of(I, C),  instance\_of(I, Y). \label{q2_1}\\
out\_value(R, X, Y) & \leftarrow & get\_value(I, C),   
	 value(R, X, Y),  relation(R),  \label{q2_2} \\
	 && term\_of(X, I), term\_of(Y, I). \nonumber
\end{eqnarray}
}
where $term\_of(X, I)$ defines a term ($X$) that is 
constructable from Skolem functions and an individual ($I$), 
$out\_member\_of(d)$ indicates that $i$ 
is an instance of the class $d$ (i.e., $d \in C(c)$),  and 
$out\_value(r, x, y)$ says that $KB(D) \models value(r, x, y)$.
This answer is correct but may contain {\em too much} information 
for users of an OOKB who have knowledge about the class hierarchy.
This is because the above description %%%of a class $c$ 
could also include values that $i$ can inherit from the 
superclasses of $c$. This can be seen in the next example.  
\begin{example} \label{ex1}
{\rm 
Let us consider the class {\small \tt Eukaryotic cell}. The description of this
class contains 88 statements of the form \eqref{eqslot1}---\eqref{eqslot2} that 
involve 167 classes and 150 equality specifications. A first-level answer\footnote{ 
   Current solvers can only approximate the answer due to 
   the infiniteness of the grounding program. We computed the answer 
   by limiting the maximum nesting level for complex terms of the term  
   to be 1 (e.g., the option {\small {\tt maxnesting}} 
   in {\bf dlv}). 
 } 
computed using \eqref{q2_1}--\eqref{q2_2}  contains  
9 atoms of the form $out\_member\_of(x)$ which indicate that
a {\small \tt eukaryotic cell} is also a {\small \tt cell}, a {\small \tt living entity}, 
a {\small \tt physical object}, etc. 
In addition, there are 643 atoms of the form $out\_value(r, x, y)$ 
which contains inverse, composition, sub-relation, and the relation value 
defined in statements of the form \eqref{eqslot1}---\eqref{eqslot2}  
and those that are obtained by the rules \eqref{p5}--\eqref{p7}. 
}
\end{example} 

The example highlights two challenges in computing the description of
an individual. First, since the grounding of the KB is infinite, it
raises the question of what counts as an adequate grounding that
returns a sufficient description of an individuals? Second, for
practical query answering applications that use \kbbio, one must
post-process the results to deciding which subset of the answers
should be preesnted to the user.  It should be noted that because of
the infiniteness of the grounded KB, current ASP solvers can be used
to approximate the answers, by setting depth bounds. Whether this will
result in acceptable performance, both in terms of the quality of the
answers and the efficiency, is a topic open for future research.

\subsection{Comparing between Classes ($\mathbf{ Q_3}$)} 

A comparison query takes the general form of ``What are the
differences/similarities between $c_1$ and $c_2$?'' (e.g., ``what are the
differences between {\em chromosome} and {\em ribosome}?'').  More
specific versions of the query may ask for specific kinds of
differences, e.g., structural differences.

The query can be represented and answered by ({\em i}) introducing 
two new constants $i_1$ and $i_2$ which are instances of $c_1$
and $c_2$, respectively; and ({\em ii}) identifying the differences
and similarities presented in the descriptions of $i_1$ and $i_2$. 
We therefore encode $I(Q_3)$ using  
the following program: 
{ 
\begin{align} 
 instance\_of(i_1, c_1).   \quad 
 instance\_of(i_2, c_2).    \quad
 comparison(i_1, c_1, i_2, c_2).    
\end{align} 
}
Let us first discuss the features that can be used in comparing individuals of two classes. 
Individuals from two classes can be distinguished from each other using  
different dimensions, either by their superclass relationship or by the relations 
defined for each class. More specifically,  they can be differentiated from each other by the
generalitation and/or specialitation between classes; or 
the properties of instances belonging to them. We will refer to these two dimensions as {\em class-dimension} 
and {\em instance-dimension}, respectively.   
We therefore define the following notions, given an answer 
set $M$ of $KB(Q_3)$: 
\begin{list}{$\bullet$}{\itemsep=0pt \topsep=0pt \parsep=0pt} 
\item {\em The set of similar classes between $c_1$ and $c_2$}: is 
the intersection between the set of superclasses of $c_1$ and of $c_2$
$ U(c_1,c_2) =  C(c_1) \cap C(c_2)$. 

\item {\em The set of different classes between $c_1$ and $c_2$}: is
the set difference between the set of superclasses of $c_1$ and of $c_2$ 
$D(c_1,c_2) = (C(c_1) \setminus C(c_2)) \cup  
 (C(c_2) \setminus C(c_1))$.
\end{list}  
where $C(c)$ denotes the set of superclasses of $c$.
 
We next discuss the question of what should be considered as a similar  
and/or different property between individuals of two different classes. 
Our formalization is motivated from the typical answers to this type 
of question such as an answer ``a chromosome has a part as protein 
but a ribosome does not'' to the query ``what is the different between 
a chromosome and a ribosome?'' This answer indicates that for each 
chromosome $x$ there exists a part of $x$, say $f(x)$, which is 
a protein, i.e., $value(has\_part, x, f(x))$ and $instance\_of(f(x), protein)$ 
hold; furthermore, no part of a ribosome, say $y$, is a protein, i.e., 
there exists no $g$ such that $value(has\_part, y, g(y))$ 
and $instance\_of(g(y), protein)$ hold. 

For a set of literals $M$ of $KB(Q_3)$ and a class $c$ with 
$instance\_of(i, c) \in M$, let $T(i, c)$ be the set of triples 
$(r, p, q)$ such that  $(r, x, y) \in \Sigma(i,c,M)$,  
$instance\_of(x, p) \in M$,  and 
$instance\_of(y, q) \in M$. $p$ ($q$) is called the domain (range) of 
$r$ if $(r, p, q) \in T(i, c)$. We define 
\begin{list}{$\bullet$}{\itemsep=0pt \topsep=0pt \parsep=0pt} 
\item {\em The set of similar relations between $c_1$ and $c_2$}: is the set 
$R^s(c_1,c_2)$ of relations $s$ such that 
({\em i}) $c_1$ and $c_2$ are domain of $s$; 
({\em ii}) $c_1$ and $c_2$ are range of $s$; or 
({\em iii}) there exist $(p,q)$ such that  $(s,p,q) \in T(i_1,c_1) \cap T(i_2,c_2)$.
\item  {\em The set of different relations between $c_1$ and $c_2$}: is the 
set $R^d(c_1,c_2)$ of relations $s$ such that ({\em i}) $c_1$ 
is  and $c_2$ is not a domain of $s$ or vice versa; 
({\em ii}) $c_1$ is and $c_2$ is not a  range of $s$ vice versa; or 
({\em iii}) there exist $(p,q)$ such that  $(s,p,q) \in (T(i_1,c_1) \setminus T(i_2,c_2)) \cup 
(T(i_2,c_2) \setminus T(i_1,c_1))$.
\end{list} 
An answer to $\mathbf{Q_3}$ must contain information from 
$U(c_1,c_2)$, $D(c_1, c_2)$, $R^s(c_1,c_2)$, and $R^d(c_1,c_2)$. 
Computing $U(c_1,c_2)$ and $D(c_1, c_2)$ rely on the rules 
for determining the most specific classes among a group of classes
which can easily be implemented using the naf-operator. 

We now describe the set of rules $R(Q_3)$, dividing it 
into different groups. First, the set of rules for computing $U(c_1, c_2)$
is as follows: 
{%\small
\begin{eqnarray}
shared(C, P, Q) & \leftarrow & comparison(X, P, Y, Q),     subclass\_of(P,C), subclass\_of(Q,C).   \quad \quad  \quad
\end{eqnarray} 
}
The rule identifies the classes that are superclass of 
both $c_1$ and $c_2$. 
We can show that $KB(Q_3) \models shared(c, c_1, c_2)$ 
iff $c \in U(c_1, c_2)$. 

The next set of rules is for computing $D(c_1, c_2)$. 
{\small
\begin{eqnarray} 
dist(C, P, Q) & \leftarrow & comparison(X, P, Y, Q), 
subclass\_of(P,C),  \naf subclass\_of(Q,C).\quad \quad  \label{diff-class-1} \\
%& & nonumber \\
%
dist(C, P, Q) & \leftarrow & comparison(X, P, Y, Q),  
 \naf subclass\_of(P,C),  subclass\_of(Q,C).   \label{diff-class-2}
\end{eqnarray} 
}
The two rules identify the classes that are superclass of 
$c_1$ but not $c_2$ and vice versa. Again,  
we can show that $KB(Q_3) \models dist(c, c_1, c_2)$ 
iff $c \in D(c_1, c_2)$. 

For computing $R^s(c_1,c_2)$ and $R^d(c_1,c_2)$, we 
need to  compute the sets $T(i_1,c_1)$ and $T(i_2, c_2)$. 
For this purpose, we define two predicates $t_1$ and $t_2$ 
such that for every answer set $M$ of $KB(Q_3)$, 
$t_k(s, p, q) \in M$ iff $(s, p, q) \in T(i_k,c_k)$ 
for $k=1,2$. Before we present the rules, 
let us denote a predicate $msc\_of$, 
called the {\em most specific class of an individual}, by the following
rules. 
{\small
\begin{eqnarray} 
not\_msc\_of(X, P) & \leftarrow & subclass\_of(Q, P), 
    instance\_of(X, P), instance\_of(X, Q). \quad \\  
msc\_of(X, P) & \leftarrow & instance\_of(X, P), 
 \naf not\_msc\_of(X, P). 
\end{eqnarray} 
}
These rules state that the class $p$ is the most specific class of 
an individual $x$ if $x$ is a member of $p$ and $x$ is not an 
instance of any subclass $q$ of $p$.  
This will allow us to define the set $T(i_1,c_1)$ and $T(i_2, c_2)$
as follows. 
{\small
\begin{eqnarray} 
3\{t_1(R, P, Q), & \leftarrow & comparison(X_1, C_1, Y_1, C_2), value(R, X, Y), \\
q\_d(R, P), && term\_of(Y, X_1), term\_of(X, X_1), \nonumber \\   
 q\_r(R, Q)\}   && msc\_of(X, P), msc\_of(Y, Q). \nonumber \\
3\{t_2(R, P, Q),    & \leftarrow & comparison(X_1, C_1, Y_1, C_2), value(R, X, Y), \\ 
q\_d(R, P),&&  term\_of(X, Y_1), term\_of(Y, Y_1), \noindent \\  
q\_r(R, Q)\}      & & msc\_of(X, P), msc\_of(Y, Q). \nonumber 
\end{eqnarray} 
}
The following rules identify relations that are similar 
between $c_1$ and $c_2$: 
{\small
\begin{eqnarray} 
shared\_property(R) & \leftarrow & comparison(X_1, C_1, Y_1, C_2), 
     t_1(R, C_1, Q_1), t_2(R, C_2, Q_2).\quad \\
shared\_property(S) & \leftarrow & comparison(X_1, C_1, Y_1, C_2), 
     t_1(R, P_1, C_1), t_2(R, P_2, C_2).  \\
shared\_property(S) & \leftarrow & comparison(X_1, C_1, Y_1, C_2), 
     t_1(R, P, Q), t_2(R, P, Q). 
\end{eqnarray} 
}
The rules say that individuals $i_1$ and $i_2$ 
from class $c_1$ and $c_2$ respectively 
share a relation $r$. The first rule says that $i_k$ ($k=1,2$) 
is a source in the relation $r$ (i.e., there exists
some $t_k$ such that $(r, i_k, t_k) \in \Sigma(i_k, c_k, M)$);  
The second rule says that $i_k$ is a destination in the relation $r$ 
(i.e., the first rule: there exists some $t_k$ such that $(r, t_k, i_k) \in \Sigma(i_k, c_k, M)$). The third rule says that there exist
some pair $t^1_k, t^2_k$ such that $t_k^1$ and $t_k^2$ are instances
of the same class and $(r, t^1_k, t^2_k) \in \Sigma(i_k, c_k, M)$. 

The set of rules for computing $R^d(c_1,c_2)$  is similar to the above set 
of rules. It is omitted here for space reason.

The key challenge in computing the differences/similarities between
classes in \kbbio\ are the same as for $Q_2$. First, since the
grounded program is infinite, one has to determine what is an adequate
description that should be used for the purposes of
comparsion. Second, even though the computation will return all
differences and similarties, the users are frequently interested in
knowing about salient differences.  The current AURA system uses a
complex set of heuristics to post process the results to group and
rank the results to draw out the salience. The description of such
heuristics is outside the scope of the present paper.

%Similar to the class differences and similarities, 
%the sets $R^s(c_1,c_2)$ and $R^d(c_1,c_2)$ can be reduced to the 
%set of most specific relations. For brevity, we omit the precise 
%definition of this notion. 
 
\subsection{Relationship between Individuals ($\mathbf{Q_4}$)} 

A relationship query takes the general form of ``What is the
relationship between individual $i_1$ and individual $i_2$?'', e.g.,  
``what is the relationship between a {\small \tt biomembrane} and a
{\small \tt carbohydrate}''?  Since this type of query refers to a path between two
individuals, it can involve significant search in the KB
making it especially suitable for solution by ASP solvers.  In more
specific forms of this query, the choice of relationships can be
limited to a specific subset of relationships in the KB. For example,
``What is the structural or functional 
relationship between individual $i_1$ and
individual $i_2$?'' We can formulate this query as follows.
  
Given a set of literals $M$ of an OOKB and a set of relations
$S$, a sequence of classes 
alternated with relation 
$\omega = (c_1, s_1, c_2, s_2, \ldots, s_{n-1}, c_n)$ is 
called a {\em path between $q_1$ and $q_n$} 
with restrictive relations $S$ in $M$ if  
there exists $instance\_of(t, c_1) \in M$ and Skolem 
functions $f_1=id, f_2,\ldots,f_{n-1}$ such that 
$value(s_i, f_i(t), f_{i+1}(t)) \in M$ for $i=1,\ldots,n-1$ and  
$instance\_of(f_i(t), c_i) \in M$ for $i \ge 2$ and  
$s_i \in S$ for $1 \le i < n$.  
A query of type $\mathbf{Q_4}$ asks for a path between 
$c_1$ and $c_2$ with restrictive relations in $S$ and is
encoded by the program $I(Q_4)$: 
{%\small
\begin{align}
 instance\_of(i_1, c_1).  \quad instance\_of(i_2, c_2). \quad
p\_relation(c_1, c_2).   \quad
include(r). \: (r \in S)   
  \nonumber 
\end{align} 
}
The answer to the query should indicate paths between $c_1$ and 
$c_2$ with restrictive relations in $S$. Observe that an answer 
can be generated by ({\em i}) 
selecting some atoms of the form $value(s, x, y)$ such 
that $s \in S$; and ({\em ii}) 
 checking whether these atoms create a path from $c_1$ to $c_2$. 
We next present the set of rules $R(Q_4)$, dividing them 
into two groups that implement the steps ({\em i}) and ({\em ii}) 
as follows. 
{\small
\begin{eqnarray}
p\_segment(R, E, C, F, D) & \leftarrow &  include(R), 
   value(R, E, F), instance\_of(E, C), \\
   &&instance\_of(F, D). \nonumber\\
\{seg(S, E, C, F, D)\} & \leftarrow &  p\_segment(S, E, C, F, D). \\
& \leftarrow &   p\_relation(C_1, C_2), \{seg(\_, \_, C_1, \_, \_)\} 0.\\
& \leftarrow &   p\_relation(C_1, C_2), 2 \{seg(\_, \_, C_1, \_, \_)\}.\\
& \leftarrow &   p\_relation(C_1, C_2), \{seg(\_, \_, \_, \_, C_2)\} 0. \\
& \leftarrow &   p\_relation(C_1, C_2), 2 \{seg(\_, \_, \_, \_, C_2)\}. 
\end{eqnarray}
}
The first rule defines possible segments of the path. 
The second rule, a choice rule, picks some arbitrary segments 
to create the path. A segment is represented by 
the atom $seg(s, e, c, e', c')$ that encodes 
a relation $s$ between $e$ (an instance of class $c$) 
and  $e'$ (an instance of class $c'$). 
The rest of the rules eliminate 
combinations that do not create a path from $c_1$ to $c_2$. 
For example, the first two constraints make sure that 
there must be exactly one segment starting from $c_1$; 
the next two ensure that there must be exactly one segment that
ends at $c_2$. The next four constraints make sure
that the segments create a path. 
{%\small
\begin{eqnarray}
& \leftarrow &   p\_relation(C_1, C_2),
   seg(S, E, C, E_1, D),  
   D \ne C_2,
 \{seg(\_, E_1, D, \_, \_)\} 0. \\
& \leftarrow &   p\_relation(C_1, C_2),
   seg(S, E, C, E_1, D), 
   D \ne C_2, 
  2 \{seg(\_, E_1, D, \_, \_)\}. \\
& \leftarrow &   p\_relation(C_1, C_2),
   seg(S, E, C, E_1, D),  
    D \ne C_2, C \ne C_1, 
   \{seg(\_, \_, \_, E, C)\} 0. \\
& \leftarrow &   p\_relation(C_1, C_2),
   seg(S, E, C, E_1, D),  
   D \ne C_2, C \ne C_1, 
   2 \{seg(\_, \_, \_, E, C)\}. \quad \quad 
\end{eqnarray} 
}
Even if one could define a suitable finite grounding of \kbbio,
computing $KB(\mathbf{Q_4})$ can be exponential in the worst case.
The implementation of this query in AURA relies on a set of heuristics
and depth-bound incomplete reasoning.  E.g., one heuristic
involves first performing the search on the subclass
relationships. The existing implementation is unsatisfactory as it
misses out important relationships.  In an ideal implementation, one
would first compute all candidate paths, and then rank them based on
user provided critieria.  Computing all such paths especially at the
runtime has been infeasible in AURA so far. We hope that ASP could 
provide a solution for an efficient path computation.

\section{Discussion}  
\label{sec:discussion}

We observe that there was no use of default negation in the
axioms~\eqref{eqclass}-\eqref{unify1} that specify OOKB.  The default
negation is used in the domain independent axioms, for example, in
axiom~\eqref{lp-eq6} and in
axioms~\eqref{diff-class-1}-\eqref{diff-class-2}.  In principle, default
negation could be used in axioms~\eqref{eqslot1} or axiom~\eqref{eqslot2},
but in our practical experience in developing \kbbio\, such usage has
not been necessary.  That is primarily because while formalizing the
textbook knowledge, one typically requires classical negation. It is
only for query answering that the usage of negation becomes critical.
If one generalizes OOKB to other domains, it may well be necessary to
use default negation in the domain specification
axioms~\eqref{eqclass}-\eqref{unify1}, but we have not considered such usage in our
work so far. Since default negation is necessary to specify query
answering for OOKB, ASP provides a compelling paradigm for declarative
specification of such reasoning.

Let us also consider comparison between using ASP vs DLs for OOKB
queries presented here.  There are two key features of OOKBs that are not
directly expressible in description logics: graph-structured objects
and (in)equality statements.  Using axioms~\eqref{eqslot1} and
~\eqref{eqslot2}, it is possible to define a graph structure. It is well
known that graph structured descriptions usually lead to
undecidability in reasoning~\cite{MotikGHS09}.  In(equality)
statements as in axiom~\eqref{unify} and~\eqref{unify1}, allow us to
relate skolem functions that have been introduced as part of two
different class descriptions.  Such modeling paradigm is not supported
by DLs. Of course, the reasoning with OOKBs in full generality is
undecidable, and it is an open question whether there exist decidable
fragments of OOKB for which the reasoning is decidable~\cite{ChaudhriHMS12a}.
 
Another important difference between a DL and ASP is in handling of
constraints.  To illustrate this difference, consider a KB that has a
statement: every person has exactly two parents, and further
individuals $p_1$, $p_2$, $p_3$ and $p_4$, such that $p_2$, $p_3$ and
$p_4$ are the parents of $p_1$. With
axioms~\eqref{constr2}-\eqref{constr6}, such a KB will be inconsistent.
In contrast, most DL system will infer that either $p_2$ must be equal
to $p_3$, or $p_3$ must be equal to $p_4$, or $p_4$ must be equal to
$p_2$. The semantics of constraints in AURA conform to the semantics
captured in axioms~\eqref{constr2}-\eqref{constr6}.

Our work on formalizing the OOKB queries in ASP has been only
theoretical, and an experimental evaluation is open for future work.
Some example answers of the queries considered in Section~\ref{sec:queries}
which are produced by the Inquire system can be seen at~\cite{Overholtzer+12}.

\section{Conclusions}  

We described the contents of an OOKB knowledge base, and formulated
ASP programs for answering four classes of practically interesting
queries.  We also presented a practical OOKB, \kbbio, whose size and
necessary features make the computation of the answers to these
queries almost impossible using contemporary ASP solvers.  The
specific challenges include developing suitable grounding strategies
and dealing with potential undecidability in reasoning with an OOKB.
Given the large overlap in features supported by OOKB and DLs, the
\kbbio\ also presents a unique dataset which could be used to explore
relative tradeoffs in reasoning efficiency across these two different
paradigms.  Being a concrete OOKB, \kbbio\ presents a real challenge
for the development of ASP-solvers. This also calls for the
development of novel query answering methods with huge programs in
ASP.  We welcome engaging with both the ASP and DL research
communities so that \kbbio\ could be used as a driver for advancing
the state of the art in efficient and scalable reasoning.

\subsection*{Acknowledgment}
This work was funded by Vulcan Inc. and SRI International.

\bibliographystyle{plain} 
% \bibliography{../../bibtex/bibfile,../../bibtex/enrico,../../bibtex/bib2010,mw}

%\newpage 
%\section*{Appendix: The Set of Independent Axioms $\Pi_R$} 
%

\end{document}